\documentclass{AISB2008}
\usepackage{siunitx}
\usepackage{wrapfig}
\usepackage[usenames,dvipsnames,svgnames,table]{xcolor}
\usepackage[normalem]{ulem}
\usepackage{graphicx}
\usepackage{microtype}
\usepackage{xspace}
\usepackage{comment}
\usepackage{booktabs}
\usepackage{multirow}
\usepackage{eurosym}
\usepackage{subfigure}
\usepackage{algorithm}
\usepackage{algorithmic}
\usepackage{mathtools}
\usepackage[percent]{overpic}
\DeclarePairedDelimiter\abs{\lvert}{\rvert}

\usepackage[square,sort,comma,numbers]{natbib}
\newcommand*{\eg}{e.g.~}
\newcommand*{\ie}{i.e.~}
\newcommand*{\etal}{et al.~}

\begin{document}
\title{Help, Anyone? A User Study For Modeling Robotic Behavior To Mitigate Malfunctions With The Help Of The User}
\author{Markus Bajones \and Astrid Weiss \and Markus Vincze\institute{All authors are with the Vienna University of Technology,
Austria. \textbf{firstname.lastname@tuwien.ac.at}}}

\maketitle
\begin{abstract}
Service robots for the domestic environment are intended to autonomously provide support for their users. However, state-of-the-art robots still often get stuck in failure situations leading to breakdowns in the interaction flow from which the robot cannot recover alone. We performed a multi-user Wizard-of-Oz experiment in which we manipulated the robot's behavior in such a way that it appeared unexpected and malfunctioning, and asked participants to help the robot in order to restore the interaction flow. We examined how participants reacted to the robot's error, its subsequent request for help and how it changed their perception of the robot with respect to perceived intelligence, likability, and task contribution. As interaction scenario we used a game of building Lego models performed by user dyads. In total 38 participants interacted with the robot and helped in malfunctioning situations. We report two major findings: (1) in user dyads, the user who gave the last command followed by the user who is closer is more likely to help (2) malfunctions that can be actively fixed by the user seem not to negatively impact perceived intelligence and likability ratings. This work offers insights in how far user support can be a strategy for domestic service robots to recover from repeating malfunctions.
\end{abstract}

\section{Introduction}
\label{introduction}
Domestic service robots are intended to assist their users for a variety of tasks in everyday life, such as fetching-and-carrying objects, reminding the user of specific activities, and entertainment. 
Despite many advances in robotics, autonomously and robustly performing these interactions is still error-prone. 
Most common problems are navigation malfunctions: High localization uncertainty, unreachable goal positions, and collisions with obstacles \cite{Zhu2012,zeng2013mobile}.
Navigation errors like these generally can cause an abortion of user interaction as the robot cannot continue to perform the task requested by the user. 

Roboticists are looking for different ways to let robots overcome their sensing and actuation limitations, such as learning better policies \cite{Argall2009}, semi-autonomous decision making \cite{Shiomi2008}, and asking for human help to reduce uncertainty \cite{Fong2003,Nicolescu2003,Rosenthal2010}. 
However, all these solutions address situations where coping strategies are already incorporated in the planning. 
Recovery strategies for error situations in Human-Robot Interaction are still rarely studied. 
Little is known yet how to best provide ways for robots to mitigate mistakes. 
An on-line survey on different strategies (apologies, compensation, and options for the user) showed that these strategies have the potential to reduce the negative impact of a breakdown \cite{Lee2010}. 
But what if navigation breakdowns happen in real life and a robot offers various recovery options for the user? 
Are users willing to help? 
Will they perform the right steps? 
And, how does the malfunctioning impact the user's perception of the robot? 
Previous work already demonstrated  that social learning \cite{Asoh1997}, collaborative control \cite{Bruemmer2005} and sliding autonomy \cite{Heger2005} are beneficial concepts, as well as symbiotic \cite{Rosenthal2012} and reciprocal relationships \cite{Lammer2014}, where the robot performs a task for the human and in return the human helps the robot.
Ono \cite{Ono2000} studied the reaction and help actions of people that encountered a robot which got stuck during navigation.
Their study showed the importance to communicate the ``robot's intention'' to the user for successful interaction.
To our knowledge suitable mitigation strategies and users' willingness to help a robot in repeated interaction breakdowns, where the robot fails to perform an action it is normally capable of performing, have not been explored.

In this paper, we present a user study with human dyads performing a task together with a robot, where several navigation problems occur and the robot offers different approaches to resolve the issue. We explore behavioral reactions towards a robot which is asking for help, as well as the impact of resolved malfunctions on perceived task contribution, intelligence, and likability of the robot. Our motivation behind this is to find out whether a user's willingness to help can be predicted from the interaction and subsequently modeled in the robot behavior. Our work offers insights in how far user support can be a strategy for domestic service robots to keep the interaction alive in situations, which cannot be planned in advance.

The paper is organized as follows: Section 2 gives an overview on the background and motivation of the performed study. Section 3 presents the methodology and scenario for our experiment and section 4 the subsequent research questions and hypotheses. Next, section 5 describes the detailed experimental design and procedure and section 6 the data analysis and results. Finally, the paper closes with a proposal for the resulting behavior modeling.

\section{Background}
\label{background}

Mobile service robots in domestic environments are still only present as lightweight devices such as vacuum cleaning robots, but not as the imagined multi-functional butlers, such as Rosie the robot maid in the Jetsons. The ultimate goal for these robots to perform tasks for and with their users in an autonomous way is still not reached, as there are limitations with respect to perception, cognition, and execution. One of the biggest challenges for mobile service robots is robust autonomous navigation and localization \cite{savkin2015safe}. Typical frequent navigation problems are: (1) High localization uncertainty, (2) an unreachable goal position, and (3) collisions with obstacles.

Substantial amount of research has already been performed on different techniques for error prevention, such as learning better policies \cite{Argall2009}, shared or sliding control \cite{Shiomi2008,Heger2005}, and proactively involving a human to resolve uncertainty in decision making \cite{Fong2003,Nicolescu2003,Rosenthal2010}. Humans as helpers can have different roles, such as supervisors \cite{Fong2003, Shiomi2008}, teachers{\slash}demonstrators \cite{Argall2009,Nicolescu2003,Hood2015}, and also naive users in the close environment of the robot \cite{Weiss2010,Huttenrauch2006}. However, all of this work largely assumes that the robot proactively requests help due to an awareness of limited capabilities (\eg missing manipulator \cite{Rosenthal2010}) or missing information (\eg map knowledge of a specific destination \cite{Weiss2010}). None of these cases assume that a mistake{\slash}breakdown happened, which needs to be fixed by the user in order to restore the interaction flow.

For instance in the CoBot studies, Rosenthal and colleagues \cite{Rosenthal2012} investigate whether the robot can proactively find people to help with a task, it obviously cannot solve on its own, such as calling the elevator (as it has no manipulator). Asking for help is integrated in the planner with different strategies (identifying if help is needed, how long to wait for help, where to search for help) to enable the robot to reach its goal. In the IURO project Weiss and colleagues \cite{Weiss2015} studied the acceptance of a robot that proactively asks pedestrians for help to find its way. Asking for the way is therefore in this scenario the intended robot behavior and can be planned in advance. Proactively and reactively asking a user for help, however, have one important aspect in common: The user needs to comply with the robot's request in order to successfully perform the task. 

Even though, situations of malfunctioning cannot be planned in advance, suitable recovery strategies can and should be planned in order to keep the interaction with the user alive. An interaction abortion without offering any mitigation strategy negatively impacts the user's perception of the robot \cite{Lee2010}. It is shown that offering appropriate recovery strategies enables the user to increase the bonding towards failed services \cite{Aaker2004,Hart1990,Spreng1995}. However, it is also proven that people often become emotionally upset when there is a service breakdown, whereby they are more dissatisfied by the failure of the recovery than by the mistake itself \cite{Bitner1990}. So far little research has been done on mitigation strategies for service robots. Prior research showed that people feel a loss of control when they do not understand why a robot fails \cite{Hinds2004}. The more autonomous a robot was, the more people blamed it for failure and explaining the reason for failure led only to little improvement \cite{Kim2006}. Moreover, it was shown that people's orientation  toward  services influences which  recovery strategy works best for them. Those with a relational orientation responded best to an apology; those with a utilitarian orientation responded best to a compensation \cite{Lee2010}. 

We focus our research on malfunctioning situations happening while the user is interacting with the robot in a situation that is normally error-free and in which the robot can offer beneficial compensation strategies to keep the interaction alive with the help of the user. 
We are interested to explore users' willingness to help in a multi-user setting, with repeated failure situations in order to identify recovery strategies that can be modeled for the robot and to assess the impact on the user's perception of the robot. 
Since even with perfect hardware and software service robots will make mistakes, studying and modeling mitigation and recovery strategies for Human-Robot Interaction is not only beneficial for current, but also for future systems.

\section {Methodology and Scenario}

The basic idea of our study is to manipulate the robot's behavior in such a way that it appears unexpected and malfunctioning and to examine how users react to different recovery strategies. We use a collaborative game as interaction scenario for a user dyad, namely building Lego models.
The overall motivation of this study is to explore under controlled conditions if people are willing to help the robot in repeated situations of malfunction. Therefore, we defined three different types of repeated unexpected navigation behaviors: high localization uncertainty, unreachable goal pose, and obstacle collision. For more details see section \ref{robbehav}.

The study was set up as a mixed design consisting of a between-subject condition: (1) \emph{Same Task - ST}: Both participants contribute in the same way to the task and equally often interact with the robot (2) \emph{Different Task - DT}: The two participants contribute in different ways to the task and subsequently one interacts more with the robot than the other. The investigation of \emph{expected vs. unexpected (malfunctioning) behavior} and the related impact of perceived task contribution, intelligence, and likability were carried out as within-design.

\begin{figure}[htbp]
\begin{center}
\includegraphics[width=0.475\textwidth]{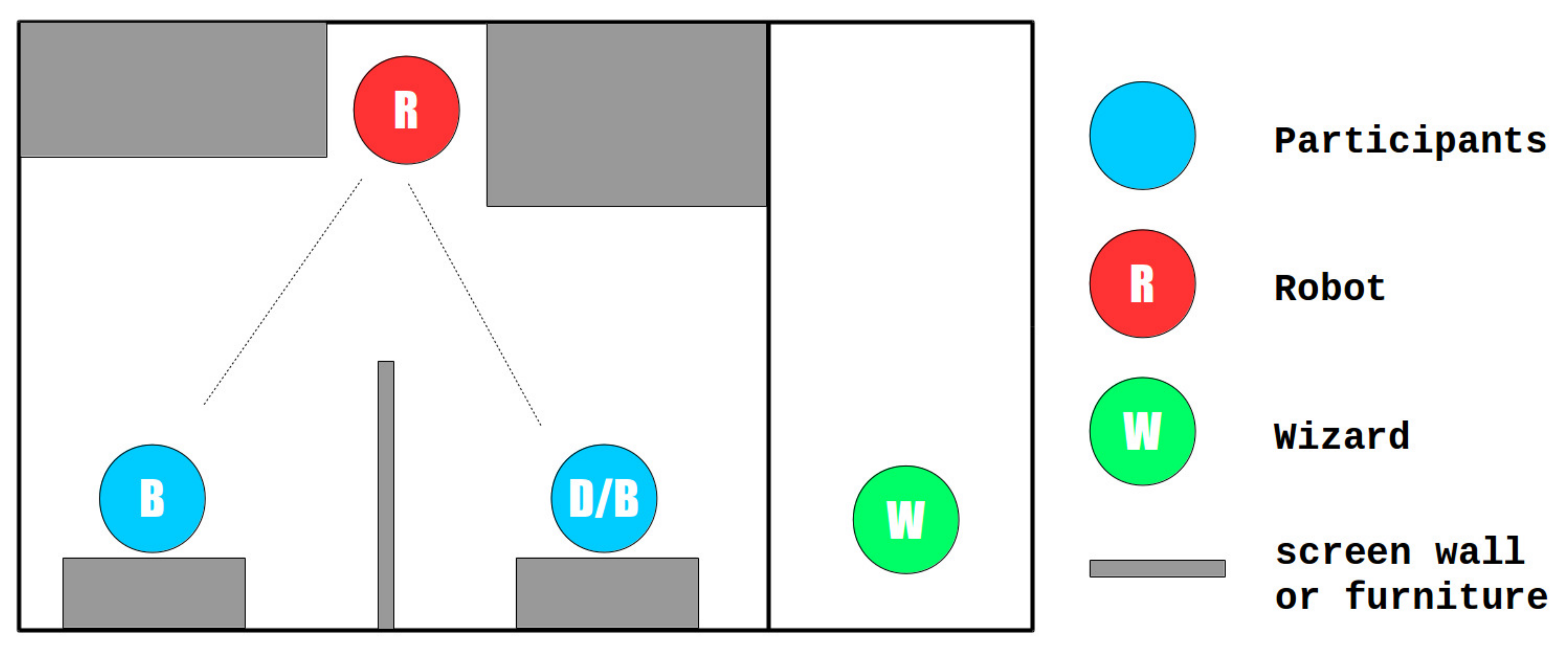} 
\caption{Experimental room layout with both participants (director{\slash}builder) at their workplaces and the robot in its docking station. Two lines indicate the direct paths between the docking station and the users' workplaces.}
\label{fig:room_setup}
\end{center}
\end{figure}

\begin{description}
\item [Different task]
Participant 1 is the director, participant 2 is the builder. The task is that participants, having different bricks at their disposal, build a Lego model together, while one participant, the director, gets instructed by the robot how the model should look like, and the builder needs to rebuild the model. The workspaces of the director and the builder are separated (Figure \ref{fig:room_setup}), so when the builder is missing a needed brick, the director can send the robot to the builder with that Lego block.
\item [Same task]
The participants have the same role, they are both builders. They both have to build a unique model shown to them by the robot, but they need to exchange bricks via the robot in order to succeed.
\item [Expected vs.\ unexpected behavior]
In both conditions (\emph{ST} and \emph{DT}) participants are asked to build in total three Lego models. 
After several correct runs, we assume that participants expect consistent behavior, as while participants are building the first Lego model no unexpected behavior emerges. 
When the robot then, during the building phase of the second model, misbehaves, participants are likely to be surprised. 
By avoiding unexpected behavior at the very beginning and very end of the interaction, we assure that the failures do not negatively impact the overall system trust \cite{Desai2013}.
\end{description}
Both participants have their own workspace separated by a screen wall and can call the robot by pressing a button.

\subsection{Manipulation of Robot Behavior}
\label{robbehav}
The behavior of the robot followed a pre-defined script that relies on three basic elements (1) linear and rotational movement, (2) pan and tilt head-movement to give the participants the notion of awareness, and (3) response to the user's button presses to call the robot to their workspace. 
The expected behaviors are:
\begin{description}
\item [Approach the user] The robot moves from the charging station to the user who pressed the call button. 
\item [Present instruction to the user] The robot shows the user how the model should look like.
\item [Navigate to the docking station] The robot will move away from the user and returns into the docking station.
\end {description}
The unexpected behavior condition introduces three common navigational failures and supplies the user with the corresponding help messages:
\begin{description}
\item[High localization uncertainty] Self-localization faces the risk of accumulating uncertainty of the estimated position of the robot over time. This increases especially during navigation with little or no landmarks (\eg in long corridors). Reaching a pre-set limit of uncertainty should stop any movement attempts as the risk of collision would be too high. In this situation the robot asked the participant: \emph{I am lost. Can somebody please push me back into my docking station?}
\item[Goal pose not reachable] Especially with the risk of a human moving into the area surrounding the goal pose it cannot be guaranteed that the robot is able to reach the desired position. In such a case the robot should stop at a pose that is balanced for small distances to the goal and offers enough free space for a human to move away from the robot (\eg not directly in front of a chair in which a person is sitting). In this situation the robot asked the participant: \emph{I was expecting to find you here. You can just push me into the right direction.}
\item[Collision with humans or obstacles] As not all robots are equipped with sensor systems that provide a \SI{360}{\degree} field of view there is always a risk of touching an obstacle during rotation or linear movement. In this case the robot needs to stop any further attempt to continue any movement. In this situation the robot told the participant: \emph{I'm stuck. Please push me away from the obstacle.}
\end{description}

\section{Research Questions}
\label{rq}

It is known that user help is beneficial to support robots in situations they cannot solve on their own \cite{Rosenthal2012}. Weiss and colleagues \cite{Weiss2010} showed that users are willing to help a robot in need; but there was no shared task between the user and the robot. For the pedestrians it was irrelevant if they could successfully help the robot. There was no multi-user setting and above all no repeated  requests for help. 
However, laboratory trials revealed that users positively experience it if they can successfully help a robot one time \cite{Lammer2014}.
There are still unexplored aspects in helping situations in HRI that we address with our study. Our two guiding research questions were:
\begin{description}
\item[Willingness to Help] Are users repeatedly willing to help the same robot?
\item[Multi-User Setting] In a multi-user setting, who is going to help? 
\end{description}
Many Human-Robot Interaction scenarios, in which the robot is asking a human for help, assume that the robot needs help at its current location and only needs help once. In most cases it is assumed that user interaction just happened, \eg the user was interacting with the robot through a user interface \cite{Shiomi2008}, a mobile device \cite{Fong2003} or via speech \cite{Lee2010,Weiss2010}. In our scenario, helping situations are repeatedly requested while the robot is navigating between two users. Therefore we aim to identify which user is helping the robot depending on the spatial and interactional situation and if repeated help is provided.
The answers to these questions are essential, if we want to be able to predict who the robot should engage first in a malfunction situation.
Moreover, we set up the following hypotheses with respect to our experimental conditions.
\begin{description}
\item[Hypothesis 1: Expected vs. Unexpected Behavior] Participants perceive a robot that shows malfunction as less intelligent, less likable, and assign a lower task contribution to it compared to a robot which always behaves correctly (within-subject variables).
\item[Hypothesis 2: Different vs. Same Task] Participants in the role of the ``director'' in the DT condition will show more helping behavior than the ``builder'', as the director has more control over the task, feels more in charge, and is more often in contact with the robot. In the ST condition no differences between the participants should be observable as they are contributing to the task in the same way (between-subject variables).
\end{description}

It is known that the task context \cite{Joose2014} and different user roles{\slash}personalities \cite{Weiss2012} can have an impact on the user's perception of a robot. So potentially it could also affect the users' helping behavior.

\section{Experiment}
\label{experiments}
To conduct our experiment we used a mobile robotic platform HOBBIT, that stands \SI{140}{\centi\meter} tall. HOBBIT is equipped with a 7-degrees of freedom arm on its right side, and uses two ASUS Xtion Pro depth sensors for self-localization, human and obstacle detection as well as gesture recognition. The robot has two actuated wheels and one caster wheel for stable movement. 
\begin{figure}[htbp]
\centering
\includegraphics[width=0.2\textwidth]{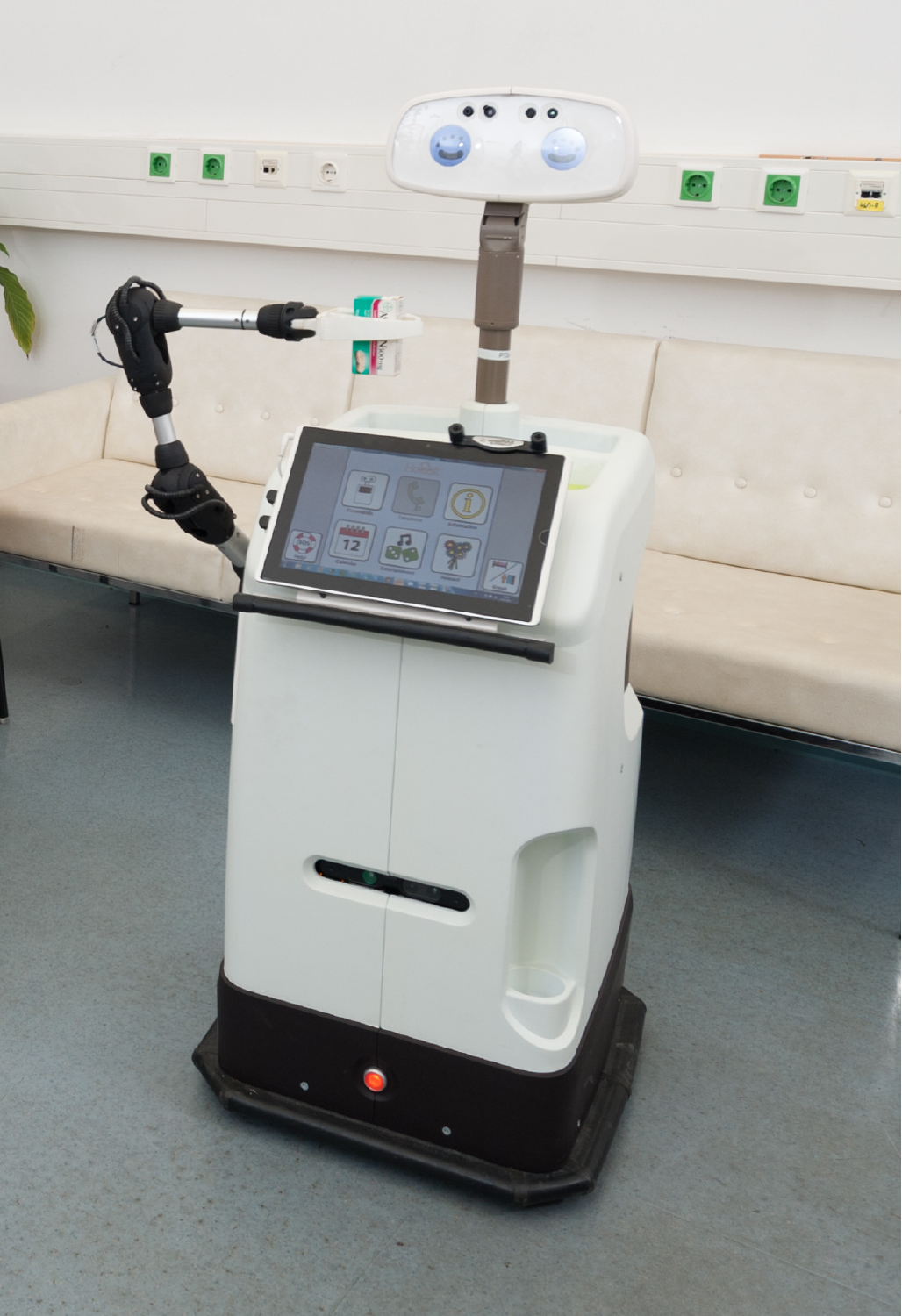} 
\caption{Hobbit - Prototype 2. Originally designed for long-term studies in which it assisted elderly through the means of fall-prevention and -detection  \cite{Pripfl2016}}
\end{figure}
The robot was controlled with a Logitech F710 wireless game pad by a wizard, seated in the adjacent room, during the experiment. In the experiment room we placed four cameras, two of which were mounted above the participants' workspace and the others on opposing walls. The wall mounted cameras were mainly used for navigating the robot. Apart from this we used a folding screen to increase the path length for the robot and Lego bricks as building blocks of the models. To measure the distances between the participants and the robot from the recorded video data we installed markings on the floor of the room (\SI{0.1}{\meter} resolution).

\subsection{Experimental Procedure}
During the introduction of the study and the questionnaire{\slash}interview rounds one experimenter was present in the room with the participants. The second experimenter was in the adjacent room to wizard the robot during the entire duration of the study.
\begin{description}

\item[Introduction and pre-interview]
The experimenter introduced the study concept and the fact that the study was recorded with multiple video cameras, and asked to sign informed consent. Then, participants were briefly interviewed on their pre-experience with robots and their expectations. 
At the end of the interview we introduced the robot, explained that it is there to transport bricks for them and that we test its navigation robustness. 
The participants were informed that the robot's behavior is completely autonomous and a short demonstration of the robot reacting to the call buttons on each person's workspace was given. 
The participants were told that the goal of the study was to test navigation in a narrow, apartment-like environment and that the robot would wait for no more than 30 seconds for a user input when needed. 
If none was received, the robot would move back into its charging station.
Before the next step, the experimenter moved the docking station and the robot to its designated place in the room (Figure \ref{fig:room_setup}) while being watched by both participants. 
Thereby it was demonstrated to the participants without explanation that the robot can be moved by pushing it. 
Moreover, it was necessary to achieve the same distance between the robot and both workspaces.

\item[In-between and post-interviews]
The actual experiment was split into three phases of interaction, explained in detail in \ref{subsub:db} and  \ref{subsub:bb}. After each phase self-reporting data was captured using a \emph{task contribution questionnaire} asking the following three questions: How much did you contribute to the task? 
How much did your counterpart contribute to the task? 
How much did the robot contribute to the task? 
All three questions had to be rated on a 7-point Likert scale. 
Additionally, participants were asked to fill in the scales \emph{perceived intelligence} and \emph{likability} from the \emph{Godspeed Questionnaire Series} \cite{Bartneck2009} and to answer three open-ended questions: Did the robot approach you every time you pressed the button? 
What worked well, what did not work? Did something unexpected happen? 
After the last interaction phase, participants were asked (additionally to the questionnaires) two open-ended questions: Would you like to have the robot at home? Imagine you were on vacation for two weeks, would you let the robot navigate on its own? 
In the end we thanked the participants and explained the setup including the fact that the robot had been remote controlled during the whole study.
\end{description}

\subsubsection{Different task (1 builder, 1 director)}
\label{subsub:db}
Builder and director are in the same room, at different desks, physically separated by a screen wall. 
A button to call the robot to the participant is placed at both workspaces and the robot is located in its charging station. 
The builder has a set of Lego bricks in a box on the desk to build a 4-level model.
\begin{description}
\item[First interaction phase (10 min)]

The director calls the robot with the button, it moves out of the docking station, to the director and shows the first level of the model on its display. The director then instructs the user how and with which bricks the model has to be put together. 
To force as many navigational movements as possible, the robot is driving back to its charging station after every delivery of the building instructions to the director.
As one of the blocks that is needed to build an exact replica of the model is not available to the builder, the director needs to be asked to provide this missing piece.
To that end the director puts this block on the robot's tray, before the builder calls the robot. 
The next step is similar to the beginning of the scenario. The director calls the robot by pressing the button and after its arrival the instructions for the last level of bricks are shown. 
The robot moves back to its charging station and waits there until the builder calls for the robot for the final step. 
The builder puts the model on the robot's tray and as final step sends it to the director for verification. 
After comparing the two models the director gives a signal to the experimenter that they are finished.
\item[Second interaction phase (20 min)]
During this phase every second motion of the robot is manipulated in such a way that one of the defined problem situations will occur. The robot announces the nature of the one of the problems (see \ref{robbehav}) via speech output and text on the screen that arose and how the users are able to resolve it. To be able to continue the given tasks one or both participants need to follow the instructions given by the robot.
\item[Third interaction phase (10 min)]
This phase is exactly the same as the first interaction phase. Every movement is performed without any problems.	
\end{description}

\subsubsection{Same task (2 builders)}
\label{subsub:bb}
At the beginning of the scenario the robot is located in its docking station and the builders are at their desks. Every participant needs to build a model, consisting of four differently colored Lego bricks. Every builder has a set of bricks in two colors, which are complementary distributed. This drives the need for each participant to collaborate with the other participant by using the robot for transporting Lego bricks.

\begin{description}
\item[First interaction phase (10 min)]
The first builder calls for the robot, which moves to the user and presents the next step to build the model. After the participant acknowledged the information or after a 30 seconds timeout the robot drives back to its charging station. 
This is repeated for the other builder as well. 
If a builder is missing a certain block, the other one can be asked to have the robot bring this part the next time. 
Whenever the robot is near the builder's workspace, one or multiple bricks can be transported by the robot.
This process is repeated until both builders complete their models, and both give a signal to the experimenter that they are finished.

\item[Second and Third interaction phase]
The behavior of the robot in both phases follows the principle explained for director and builder in \ref{subsub:db}
\end{description}

\subsection{Participants}
In total 38 participants took part in this study in 19 mixed-gender dyads, whereby ten dyads were assigned to the \emph{Same Task - ST} condition and nine to the \emph{Different Task - DT} condition. 
The average age of participants was 34.67 (SD 11.98) in the \emph{ST} condition and in the \emph{DT} condition 25.78 (SD 7.78). 
In the \emph{DT} study condition, one pair knew each other beforehand as they were study colleagues. 
Even though we recruited our participants over public marketplace websites, all of our participants showed an educational level above average and 75\% were students. Seven \emph{DT} and three \emph{ST} participants claimed that they were already in contact with robots before the study, involving entertainment and toy robots, vacuum cleaning robots, and industrial robots. All others stated that they were complete novices to the topic.
Participants received monetary compensation for taking part in the study.

\section{Analysis and Results}
In this section we first focus on the type of data we collected, continue with the evaluation of it to seek support for our hypotheses. Based on those results we give answers to our driving research questions.  

We have obtained three main types of data: (1) behavioral data, (2) verbal statements during the experiment, and (3) self-reporting data. 
The first two were captured in video recordings, in which we annotated (1) the number of errors and of the participants helping the robot, (2) the fact who of the users gave the last command before the robot's failure happened, (3) the distance between both participants and the robot in the moment in which the robot asked for help, as well as (4) the time between the robot asking for help and receiving help. Further we transcribed the user dialogues.
Additionally, we analyzed the data from the questionnaires as well as the \emph{perceived intelligence} and \emph{likability} scales from the \emph{Godspeed Questionnaire Series} and the open-ended questions were thematically categorized.

Out of the 20 planned runs one had to be aborted during the first phase, as the robot's breaks got activated and the ability to move could not be restored in time to continue the trial. Thus we could not use any data from this dyad. During three other trials the robot experienced similar issues, of which two occurred after the fifth and one after the sixth failure situation due to technical failures in the navigation software. The data from those trials were analyzed as the issues did not have an impact on the already given assistance by the participants and the problem was resolved before the third interaction phase.

In both conditions all participants mentioned in the self-reporting that the interaction in the second interaction phase did not work out as smoothly as in the first and third phase. Participants stated that the robot behaved unexpectedly and needed help. Only three participants considered the robot's behavior just as unexpected, but not malfunctioning and five participants considered it as malfunctioning, but not unexpected. In other words, five participants acknowledged that malfunctions are not necessarily unexpected and three did not even consider the robot's behavior as malfunction, because the robot could recover with the help of the user.

\subsection{H1: Expected vs. Unexpected Behavior}
The self-reporting data on the task contribution (self, other, robot), perceived intelligence, and likability (see Table \ref{tab:result3}) revealed the expected tendencies with respect to \emph{Hypothesis 1: Participants perceive a robot that shows malfunction as less intelligent, less likable, and assign a lower task contribution to it compared to a robot which always behaves correctly (within-subject variables)}. 
The \emph{perceived task contribution of self and other} increased in both conditions for the second interaction phase as expected; similarly the \emph{robot contribution} was rated lower in the second phase compared to phase one and three. Repeated measures ANOVAs did not reveal that these differences were statistically significant over the three measurement points in time for the any of the three categories, however, this is most likely due to the small sample size.
Similar tendencies could be found for the results on perceived intelligence and likability. 
In the second interaction phase in which the robot showed unexpected behaviors the ratings for \emph{perceived intelligence} were lower in both conditions. 
For the \emph{likability} rating this was only true for the \emph{DT} condition. However, these small differences were also not statistically significant.

This also indicates that even frequent malfunctioning of the robot, as it happened in the second interaction phase, does not heavily impact the robot's perceived intelligence and likability as well as the overall experience, which could be due to the fact that the robot offered recovering strategies, which in the end enabled it to fulfill its task. Therefore, it was still considered intelligent and also likable. 
Moreover, similar as in the study of \cite{Fink2014}, participant's behavioral reactions showed that the monotonous task of building objects was more engaging in the beginning because of the need to help the robot (an effect that was especially observable in the \emph{DT} condition). 
However, it was also noticeable that the frequent demand for help was soon becoming annoying for the participants, especially in cases where it was necessary to help the robot during its very last navigation to complete the given task.
	
\begin{table}
\centering
\caption{Descriptive results on perceived intelligence (PI), likability (LI), task contribution self ($TC_S$), task contribution other ($TC_O$), and task contribution robot ($TC_R$). 
Lower ratings could be observed for PI, LI, and $TC_R$ in Phase 2.
PI and LI were extracted from 5-point, $TC_S$, $TC_O$ and $TC_R$ from 7-point Likert scales}
\label{tab:result3}
\resizebox{0.475\textwidth}{!}{%
\begin{tabular}{@{}llllllll@{}}
\toprule
       &    & \multicolumn{2}{c}{Phase 1} & \multicolumn{2}{c}{Phase 2} & \multicolumn{2}{c}{Phase 3} \\ \cmidrule(l){3-4} \cmidrule(l){5-6} \cmidrule(l){7-8}
       &    & mean    & SD    & mean    & SD    & mean    & SD    \\ \midrule
PI     & DT & 3.69     & 0.83        & 3.28     & 0.76        & 3.82     & 0.77        \\
PI     & ST & 3.77     & 0.6         & 3.43     & 0.83        & 3.83     & 0.67        \\
LI     & DT & 4.1      & 0.8         & 3.87     & 0.74        & 4.29     & 0.65        \\
LI     & ST & 4.22     & 0.75        & 4.31     & 0.88        & 4.37     & 0.74       \\
$TC_S$ & DT & 4.89     & 1.6         & 5.72     & 1.13        & 5.33     & 1.61        \\
$TC_S$ & ST & 5.35     & 1.5         & 6.25     & 0.97        & 5.4      & 1.64        \\
$TC_O$ & DT & 5.44     & 1.34        & 5.83     & 1.1         & 5.56     & 1.29        \\
$TC_O$ & ST & 5.15     & 1.66        & 6.2      & 1.36        & 5.45     & 1.64        \\
$TC_R$ & DT & 4.72     & 1.93        & 3.78     & 1.59        & 4.67     & 1.57        \\
$TC_R$ & ST & 5.1      & 1.55        & 4.05     & 1.88        & 5.55     & 1.36        \\ \bottomrule
\end{tabular}
}
\end{table}

\subsection{H2: Different vs. Same Task}
Even though we expected that there will be a difference in the amount of help actions between the same and different task conditions this could not be observed. 
In Table \ref{tab:result2} the average amount of help actions provided to the robot by the users in both conditions is provided.
Overall, no differences in the amount of helping behavior were observed for the two conditions using a Mann-Whitney U test (\emph{ST} Mdn=20.02 and \emph{DT} Mdn=18.78; U=164.50, z=-0.36, p=0.74) and also not in pairwise comparisons within the \emph{DT} condition (director Mdn= 25.44; builder Mdn=12.12; U=13.31, z=2.52, p=0.70). Moreover, the reaction times to the robot requests did not differ significantly between \emph{ST} (Mdn=51.27) and \emph{DT} (Mdn=62.12) participants, U=1262.50, z=-1.79, p=0.074 and also the time to help did not (ST Mdn=56.18, DT Mdn=56.84, U=1547.50, z=-0.11, p=0.91) as shown in Figure \ref{fig:time_to_help}.
Thus our \emph{Hypothesis 2: Participants in the role of the ``director'' in the DT condition will show more helping behavior than the ``builder'', as the director has more control over the task, feels more in charge, and is more often in contact with the robot} is rejected.

\begin{table}
\centering
\caption{Average number of help actions that the participants performed for the robot, broken down to every first and second user for both conditions \emph{ST} and \emph{DT}}
\label{tab:result2}
\resizebox{0.365\textwidth}{!}{%
\begin{tabular}{@{}llllll@{}}
\toprule
                    & mean                   & SD                     & user  & mean  & SD    \\ \midrule
\multirow{2}{*}{ST} & \multirow{2}{*}{3.2}   & \multirow{2}{*}{1.005} & $u_1$ & 2.7   & 0.949 \\
                    &                        &                        & $u_2$ & 3.7   & 0.823 \\
\multirow{2}{*}{DT} & \multirow{2}{*}{3.167} & \multirow{2}{*}{0.985} & $u_1$ & 2.667 & 1     \\
                    &                        &                        & $u_2$ & 3.667 & 0.707 \\ \bottomrule
\end{tabular}
}
\end{table}

Overall, only six of the 38 participants mentioned in their reply to the final questions that the robot could have given more concrete statements how to help, but only two of them specified that clear statements on distances would be helpful, which are obviously hard to make for real-life situations.
In the \emph{ST} condition, four participants could imagine to have the robot at home (three in the \emph{DT} condition), four could imagine to have it in a more advanced prototype stage (one in the \emph{DT} condition), and the other ten (14 in the \emph{DT} condition) could not imagine to have the robot at home at all (two participants did not answer this question). 
For most participants, the reason was that the robot did not demonstrate a useful behavior which they could need at home and not the fact that it was malfunctioning in the second interaction phase, this reason was only mentioned by two participants.
Regarding the question if people were willing to let the robot navigate autonomously in their home while they were on vacation, 5 \emph{ST} and 6 \emph{DT} participants stated a clear yes and 5 \emph{ST} and 8 \emph{DT} a clear no. 
In total, 7 \emph{ST} and 4 \emph{DT} participants said that they could only imagine this in case the navigation works more robustly or another human is present.
In both conditions (\emph{ST} and \emph{DT}), the participants provided the robot with the necessary help. Therefore we assume that the role of the user as defined in our scenario has little to no impact on the helping behavior even in situations of repeated malfunctions.
In follow-up studies we plan on focusing on \emph{ST} scenarios as they provide richer data through the higher engagement of both participants. 

\subsection{Are users repeatedly willing to help the robot?}
To give an informed answer to our first research question we counted the number of times the robot received help from one of the users. 
In the \emph{ST} condition in every failure situation at least one of the users followed the instructions of the robot to keep the interaction alive. 
Only one dyad in the \emph{DT} condition did not help after multiple attempts of the robot to get help. 
They asked the experimenter back into the room it was explained to them that they should just follow the given instructions. 
Afterwards they helped in every situation in which the robot showed unexpected behavior. 
It was further observed that even if a non-triggered failure situation came up users tried the instructions that the robot gave them during earlier error situations. 
The issue of an user accidentally pressing the bumper of the robot that activated the motor breaks, thus leaving the robot unable to move for a short time was observed multiple times during our experiments.
Table \ref{tab:result1} answers our first research question as it shows that all of our 38 users were willing to help the robot in case of repeated failure situations.

\subsection{In a multi-user setting, who is going to help?}
Furthermore, we were interested in finding patterns to explain and possibly predict who of the participants are going to help the robot. 
We analyzed how often the user that gave the last command assisted the robot during the failure situation. 
The results show that over both conditions $90.18 \%$ of the users helped the robot if they were the last person who gave the robot a command. 
Also, out of the nine cases where both participants acted to help the robot, only in one situation the last commanding user was not helping at all and in another case was not the first one of the two to help the robot. 
The data suggests that the user who gave the last command felt more responsible for the robot during the navigational task and was therefore more willing to help it. 

\begin{figure}
	\centering
	\subfigure[Same Task (ST)
	\label{subfig-1:dummy}]{%
		\begin{overpic}[width=0.47\textwidth, grid=off]{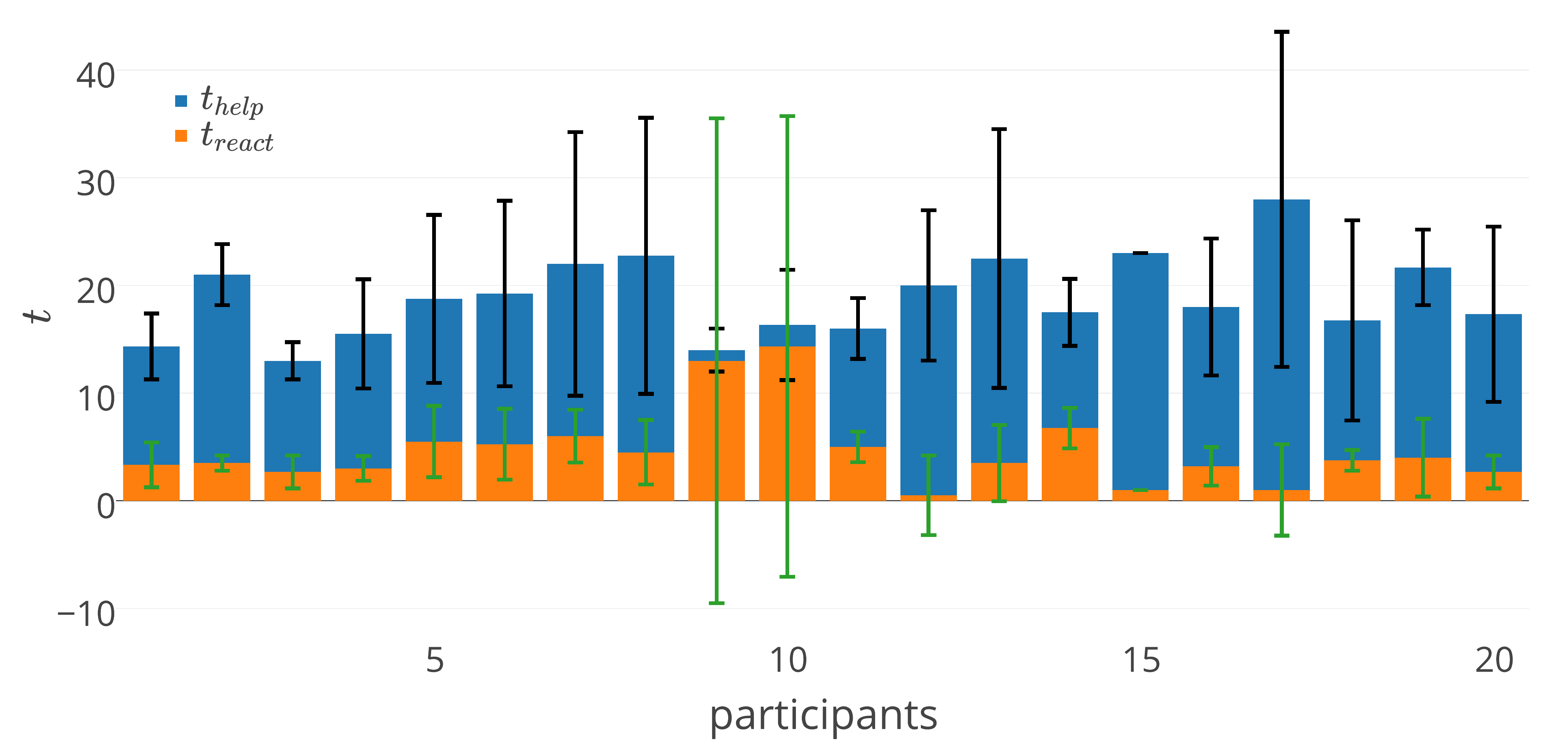}
			\put(10, 35.5){\includegraphics[scale=.07] {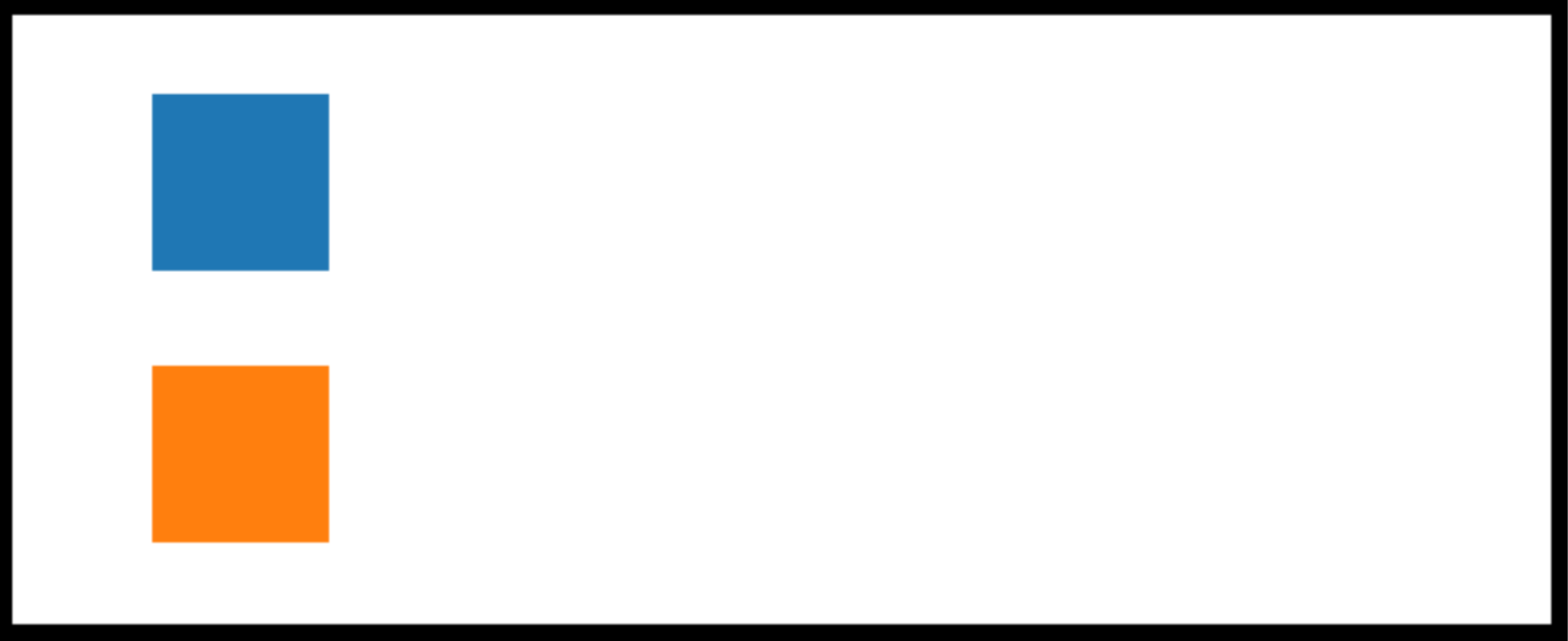}}
        		\put(16, 39.5){$t_{help}$}
        		\put(16, 36.5){$t_{react}$}
        		\put(25,5){\colorbox{white}{   }}
        		\put(49,5){\colorbox{white}{   }}
        		\put(71,5){\colorbox{white}{    }}
        		\put(94,5){\colorbox{white}{   }}
        		\put(44,1.25){\colorbox{white}{       }}
        		\put(47,1.5){\colorbox{white}{       }}
        		\put(50,1.5){\colorbox{white}{       }}
        		\put(50,1.25){\colorbox{white}{       }}
        		\put(53,1.5){\colorbox{white}{       }}
        		\put(56,1.5){\colorbox{white}{       }}
        		\put(46,0.5){\tiny{dyad}}
        		\put(8.5, 8){$\underbrace{}_1$}
        		\put(17.5, 8){$\underbrace{}_2$}
        		\put(26.5, 8){$\underbrace{}_3$}
        		\put(35.5, 8){$\underbrace{}_4$}
        		\put(44.5, 8){$\underbrace{}_5$}
        		\put(53.5, 8){$\underbrace{}_6$}
        		\put(62.5, 8){$\underbrace{}_7$}
        		\put(71.5, 8){$\underbrace{}_8$}
        		\put(80.5, 8){$\underbrace{}_9$}
        		\put(89.5, 8){$\underbrace{}_{10}$}
        		\put(0.5, 28.5){\rotatebox{90}{$[\si{\second}]$}}
      	\end{overpic}
    }
    \subfigure[Different Task (DT)
    \label{subfig-2:dummy}]{%
    		\begin{overpic}[width=0.47\textwidth, grid=off]{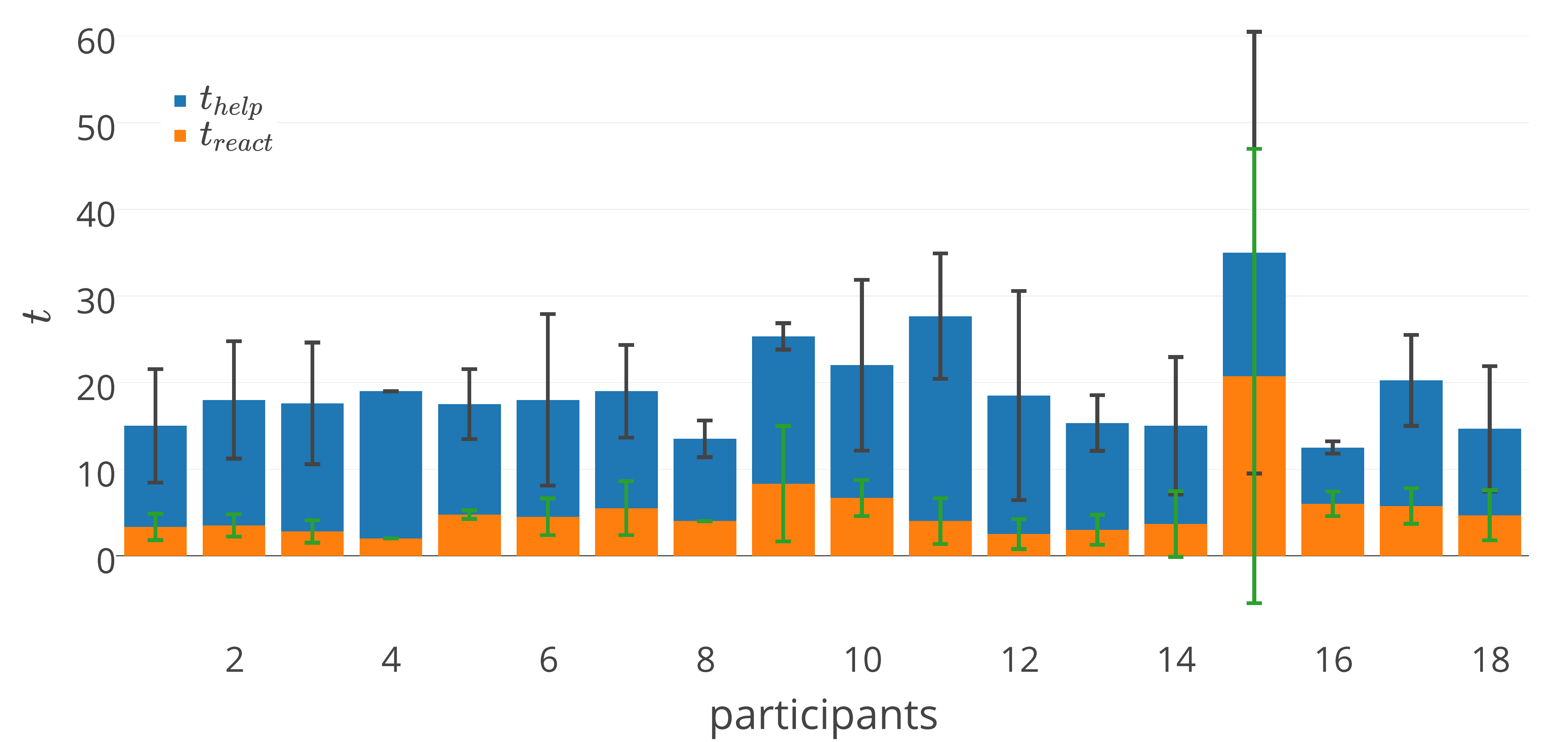}
    			\put(10, 35.5){\includegraphics[scale=.07] {legend}}
        		\put(15,39.5){$t_{help}$}
        		\put(15,36.5){$t_{react}$}
        		\put(0.5, 28.5){\rotatebox{90}{$[\si{\second}]$}}
        		\put(14,5){\colorbox{white}{   }}
        		\put(24,5){\colorbox{white}{   }}
        		\put(34,5){\colorbox{white}{    }}
        		\put(44,5){\colorbox{white}{   }}
        		\put(53,5){\colorbox{white}{   }}
        		\put(63,5){\colorbox{white}{   }}
        		\put(73,5){\colorbox{white}{    }}
        		\put(84,5){\colorbox{white}{   }}
        		\put(94,5){\colorbox{white}{   }}
        		\put(44,1.25){\colorbox{white}{       }}
        		\put(47,1.5){\colorbox{white}{       }}
        		\put(50,1.5){\colorbox{white}{       }}
        		\put(50,1.25){\colorbox{white}{       }}
        		\put(53,1.5){\colorbox{white}{       }}
        		\put(56,1.5){\colorbox{white}{       }}
        		\put(46,0.5){\tiny{dyad}}
        		\put(9.5, 8.5){$\underbrace{}_1$}
        		\put(19.5, 8.5){$\underbrace{}_2$}
        		\put(29.5, 8.5){$\underbrace{}_3$}
        		\put(39.5, 8.5){$\underbrace{}_4$}
        		\put(49.5, 8.5){$\underbrace{}_5$}
        		\put(59.5, 8.5){$\underbrace{}_6$}
        		\put(69.5, 8.5){$\underbrace{}_7$}
        		\put(79.5, 8.5){$\underbrace{}_8$}
        		\put(89.5, 8.5){$\underbrace{}_9$}
      	\end{overpic}
    }
\caption{Average duration in seconds for each single participant from the moment the robot asked for help until the participant started moving towards the robot $t_{react}$ and until the robot acknowledged that the assistance was successful $t_{help}$ during \emph{ST} (a) and \emph{DT} (b). Each pair of columns represent one dyad of the participants. The durations did not differ significantly between both conditions.}
\label{fig:time_to_help}
\end{figure}

However, this behavior could not always be observed, and is not always a good prediction for the robot. 
Thus we investigated the possibility that not the participant giving the last command, but the one who happens to simply be closer to the robot is the one who helps it more often.
The possibility that the closest user assists more often can be helpful in situations in which (1) the last command from a user is too far back in time, (2) the needed help prevents the robot to seek this user,  or (3) when the robot is acting autonomously to achieve a certain task that has not been triggered by a user. An obvious example for this is the task to move back to the charging station when the battery level of the robot drops below a threshold.
In Figure \ref{fig:time_to_help} we show $t_{help}$ as well as $t_{react}$, the time between $t_{ask}$ and the moment the first helping user started moving towards the robot $t_{move}$. Some users were observing the robot closely after the first failure so that they started moving towards the robot before it actually asked for help, which resulted in negative values for $t_{react}$ as the measurement was designed to be taken from the robot's question.
This decision was made as we were not able to guarantee the same time span between the occurrence of the failure and the robot asking for help, thus increasing potential time variations between participants based on their attention to the robot or their given task.

\begin{table}
\centering
\caption{Number of failure situations and which user helped; $u_{last}$ - the last user giving a command; $u_{closest}$ - the closer user helping; $u_{both}$ - both users helping.}
\label{tab:result1}
\resizebox{.475\textwidth}{!}{%
\begin{tabular}{@{}llllll@{}}
\toprule
                                                           & $u_{last}$ & $u_{closest}$ & $u_{both}$ & \begin{tabular}[c]{@{}l@{}}given\\ help\end{tabular} & \begin{tabular}[c]{@{}l@{}}requested\\ help\end{tabular} \\ \midrule
ST                                                         & 52      & 43      & 6      & 58                                                   & 58                                                       \\
DT                                                         & 49      & 46      & 3      & 54                                                   & 54                                                       \\
$\sum$                                                     & 101     & 89      & 9      & 112                                                  & 112                                                      \\
percentage                                                 & 90.18\% & 79.46\% & 8.04\% & 100\%                                                &                                                          \\ \bottomrule
\end{tabular}
}
\end{table}

After closely studying the video recordings we believe that the negative $t_{react}$ can be explained by the fact that those participants already finished their part of the current task and observed the robot's movement very closely and started moving towards the robot before it even finished the vocal statement that it needed assistance. This was a very positive side-effect indicating that the mitigation strategies we offered, were even predictively used by the participants in order to keep the interaction flow more smooth.

\section{Implications for Autonomous HRI}
Up to now Rosenthal \etal provide the only architecture that proactively incorporates the help from users in their planner \cite{Rosenthal2012}. 
However, their algorithm is designed to deal with known limitations of the robot, \ie the missing ability to press an elevator button. Our motivation is to develop an architecture that allows us to predict who to ask to recover from a navigation problem. Therefore, we plan on incorporating the findings of this study in a decision-making process designed for use on a mobile robot platform equipped with a laser range finder for distance measurement and a camera system for person detection and selection.
Specifically, we will use Algorithm \ref{alg:future_work} that incorporates the increased willingness to help of a user that gave the last command or is otherwise closest to the robot to select the user to ask for assistance in a not wizarded follow up study.
Work an a follow-up study to extend our initial data and to perform extensive analysis is currently ongoing.

\begin{algorithm}[htbp]
\begin{algorithmic}
\REQUIRE{failureResolved $\leftarrow$ \FALSE, firstRun $\leftarrow$ \TRUE\\ 
reactionThreshold $\leftarrow t_{react}$, 
helpTreshold $\leftarrow t_{help}$\\
users $\leftarrow$ GetAllVisibleUsers(sensorData)}
\WHILE{$\neg$failureResolved \AND $\abs{users} > 0$}
\IF{$t_{now} - t_{lastcmd} < $ commandTreshold \AND firstRun}
\STATE{user $\leftarrow$ GetLastCommandingUser()}
\STATE{firstRun $\leftarrow$ \FALSE}
\ELSE \STATE{user $\leftarrow$ GetClosestUser()} 
\ENDIF
\IF{AbleToNavigateSafely()}
\STATE{NavigateTo(user)}
\ENDIF
\STATE{PutFocusOnUser(getHeadPose(user))}
\STATE{AskForHelp(user, helpType)}
\STATE{willing $\leftarrow$ WaitForReaction(reactionThreshold)}
\IF{willing}
\STATE{failureResolved $\leftarrow$ Resolve(helpType, helpTreshold)}
\ENDIF
\IF{$\neg$failureResolved}
\STATE{users $\leftarrow$ users $-$ user}
\ENDIF
\ENDWHILE
\RETURN failureResolved
\end{algorithmic}
\caption{User selection to ask for assistance} 
\label{alg:future_work}
\end{algorithm}

\section{Conclusions}
In this work we investigated in a Wizard-of-Oz study, if user support in situations where a robot is repeatedly malfunctioning can be a beneficial mitigation strategy. 
Our data gave us empirical evidence on users' reactions and willingness to help a robot that is supposed to assist them in fulfilling a task. 
This data enables us to model ``planning for help'' in the robot behavior for situations we cannot foresee and where the robot only reactively can ask for help to keep the interaction alive. 
Moreover, the reflective self-reporting data from participants is also encouraging that even frequent malfunctioning situations do not heavily negatively impact users' perception of the robot. 
In other words, integrating mitigating strategies, such as the ones presented in this paper, into the robot's behavior coordination can bring us one step further to have autonomous service robots in domestic environments in near future; especially when we know that people are willing to accept that their robot needs a little bit of help from time to time.

\section{Acknowledgements}
\label{acknowledgements}
This work has been partially funded by the European Commission under grant agreements FP7-ICT-610532 SQUIRREL and FP7-IST-288146 HOBBIT, as well as by the Austrian Science Foundation~(FWF) under grant agreement T623-N23, V4HRC.

\bibliographystyle{AISB2008}
{\scriptsize
\bibliography{willingness_to_help_new_frontiers_camera_ready}
}

\end{document}